\documentclass[10pt,twocolumn,letterpaper]{article}

\usepackage{cvpr}
\usepackage{times}
\usepackage{graphicx}
\usepackage{amsmath}
\usepackage{amssymb}
\usepackage{verbatim}
\usepackage{multirow}
\usepackage{array}
\usepackage{color}

\usepackage[pagebackref=true,breaklinks=true,letterpaper=true,colorlinks,bookmarks=false]{hyperref}

\cvprfinalcopy 

\def\cvprPaperID{4} 

\ifcvprfinal\fi
\begin{document}

\title{DupNet: Towards Very Tiny Quantized CNN  with Improved Accuracy for Face Detection}

\author{Hongxing Gao, Wei Tao, Dongchao Wen, Junjie Liu\\
	Canon Information Technology (Beijing) Co., LTD\\
	{\tt\small \{gaohongxing,taowei,wendongchao, liujunjie\}@canon-ib.com.cn}
	\and
	Tse-Wei Chen, Kinya Osa, Masami Kato\\
	Device Technology Development Headquarters, Canon Inc.\\
	{\tt\small twchen@ieee.org}
}

\maketitle

\ifcvprfinal\thispagestyle{empty}\fi

\begin{abstract}
		
Deploying deep learning based face detectors on edge devices is a challenging task due to the limited computation resources. Even though binarizing the weights of a very tiny network gives impressive compactness on model size (\eg 240.9 KB for IFQ-Tinier-YOLO), it is not tiny enough to fit in the embedded devices with strict memory constraints. In this paper, we propose DupNet which consists of two parts. Firstly, we employ weights with duplicated channels for the weight-intensive layers to reduce the model size. Secondly, for the quantization-sensitive layers whose quantization causes notable accuracy drop, we duplicate its input feature maps. It allows us to use more weights channels for convolving more representative outputs. Based on that, we propose a very tiny face detector, DupNet-Tinier-YOLO, which is 6.5$\times$ times smaller on model size and 42.0\% less complex on computation and meanwhile achieves 2.4\% higher detection than IFQ-Tinier-YOLO. Comparing with the full precision Tiny-YOLO, our DupNet-Tinier-YOLO gives 1,694.2$\times$ and 389.9$\times$ times savings on model size and computation complexity respectively with only 4.0\% drop on detection rate (0.880 vs. 0.920). Moreover, our DupNet-Tinier-YOLO is only 36.9 KB, which is the tiniest deep face detector to our best knowledge.
	
\end{abstract}

\section{Introduction}\label{sec:intro}


Deep neural networks have demonstrated impressive accuracy in many computer vision applications such as image classification, object detection and recognition, semantics segmentation, etc. However, their increasing computation cost leads to the requirement of high-end devices such as GPU for real-time inference. It has been a challenging task to deploy the deep network based face detector on the edge devices due to their limited resources (e.g. memory size and computation power). To deploy the deep models on the edge devices, lots of approaches have been proposed, such as network pruning~\cite{Prun1,Prun2}, efficient architecture design (e.g. MobileNet~\cite{MobileNetv1,MobileNetv2}) and quantized networks~\cite{XNOR,HWGQ}. Especially, for the embedded devices, quantized networks are particularly attractive because of their impressive compression ratio (\eg 32$\times$ times savings on model size) and easy conversion for fixed-point representation. For instance, IFQ-Net~\cite{IFQNet} designs a tiny fixed-point face detector (240.9 KB) through slimming, quantizing and fixed-point converting the layers of Tiny-YOLO network~\cite{TinyYOLO}. Even though the fixed-point converting is lossless,  the accuracy drop caused by slimming and quantizing is still notable.

\begin{figure}[t]
	\begin{center}
		\includegraphics[width=\linewidth]{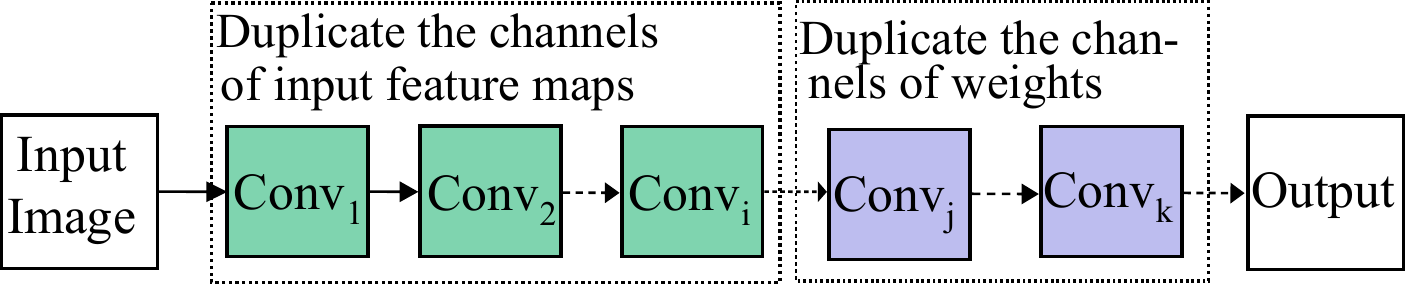}
	\end{center}
	\caption{Strategies for DupNet. We duplicate the input feature maps of the  quantization-sensitive layers which usually locate in the lower part of a network to improve the accuracy. Besides, to compress the overall model size, it employs duplicated weights for weight-intensive layers that usually locate in the upper part.}
	\label{fig:pipeline}
\end{figure}


In this paper, to compress the model size and meanwhile improve the accuracy of a quantized network, we propose DupNet as shown in Figure~\ref{fig:pipeline}. Firstly, to compress the model size, DupNet employs weights with duplicated channels for weight-intensive layers which usually are the upper layers. On the other hand, to improve the accuracy, DupNet duplicates the input feature maps and thus uses more weights channels for the quantization-sensitive layers (usually locate in the lower part of the network) whose quantization causes significant accuracy drop. In details, to further compress a quantized network, we force the weights of weight-intensive layers to have identical channels. As shown in Figure~\ref{fig:WXdup}, to generate such identical channels,  we employ template weights $\textbf{W}_t$  which has less channels than input feature maps and duplicate it to $\textbf{W}_{dup}$ for proper convolution. During inference time,  $\textbf{W}_{dup}$ can be restored from the template weights $\textbf{W}_{t}$. Consequently, model size savings can be easily achieved by only storing $\textbf{W}_{t}$.

Another major issue of the quantized network is that the accuracy is usually downgraded by a large margin. For example, in XNOR-Net~\cite{XNOR}, binarizing both weights and feature maps of AlexNet~\cite{AlexNet} leads to 12\% accuracy drop on ImageNet dataset~\cite{ImageNet}. In the case of face detection, we observe that the accuracy drop is mainly caused by the quantization of several specific layers, named as quantization-sensitive layers (see Section~\ref{subsec:exp-Xdup}). Usually, they are the lower layers which only have small amount of output feature maps. Thus, quantizing them into extremely low-bits severely harms the representative power of the output features. To address the problem, one may simply employ more feature maps or quantize them into higher bits, both of which would increase the memory usage on feature maps.

In this paper, we propose to further duplicate the input feature maps of the quantization-sensitive layers to improve its accuracy (Figure~\ref{fig:WXdup}). It is true that simply duplicating the feature maps does not introduce extra information. However, it allows us to use weights with more channels (not identical) for convolving more representative outputs. The advantage of our method is that it does not require extra memory on feature maps which is a critical issue for the embedded devices. Nevertheless, it does increase the memory usage on weights. However, as will be demonstrated in Section~\ref{subsec:exp-Xdup}, we experimentally found out that the quantization-sensitive layers are usually the lower layers of a network which only have small amount of weights. Consequently, such memory increase does not affect the network much. 

In summary, we propose DupNet which employs duplicated weights for the weight-intensive layers and duplicates the input feature maps for the quantization-sensitive layers of a quantized network. The benefits of our proposal are two-folds: 1) it reduces the model size of a quantized network by duplicated weights for weight-intensive layers; 2) it increases the accuracy through duplicating the input feature maps of its quantization-sensitive layers. Based on the DupNet, we design a very tiny quantized CNN with impressive improvement on accuracy for face detection. The model size of our network is only 36.9 KB which is the tiniest deep learning based face detector to our best knowledge.


\section{Related Work}\label{sec:RelatedWork}


\subsection{Face Detection}
Two main approaches, namely one-stage and two-stage methods,  have been successfully inherited from object detection domain for face detection. Two-stage methods follow a common two steps pipeline: 1) generates a set of region proposals with their local features; 2) pass them to a network for classifying detected objects and regressing their bounding boxes. For example, Faster-RCNN~\cite{FasterRCNN} proposes an efficient Region Proposal Network (RPN) to generate region proposals and then use Fast-RCNN network to refine the proposals. To improve the speed of Faster R-CNN,  RFCN~\cite{RFCN} proposes to share RPN network and Fast-RCNN network. In order to further improve the speed, Li~\etal~\cite{LightHeadRCNN} proposes Light Head RCNN, which employs light weight head network to reduce the computation complexity. To speedup the R-FCN network for detecting 3000 object classes, Singh~\etal ~\cite{RFCN-3000} propose to only employ position-sensitive feature maps for several predefined super-classes.

On the other hand, one-stage approaches usually employ a single network to classify and regress the objects~\cite{SSD,YOLO,YOLOv2}and thus usually can run faster. For example, YOLO~\cite{YOLO} predicts 2 bounding boxes in each of the $7\times7$ grids for VOC object detection~\cite{VOC}. Furthermore, YOLOv2~\cite{YOLOv2} employs fully convolution network that results in $m\times n$ grids (m, n are the width and height of the output feature) and uses predefined anchors to better predict the bounding boxes of the objects. In~\cite{DetNet}, Li~\etal propose a backbone network to improve the accuracy by maintaining high resolution for feature maps and reduce the computation complexity by decreasing the width of upper layers.

In spite of the enormous progresses for reducing the complexity of two-stage methods, such region proposal based frameworks may be expensive for embedded devices because they usually need to store the features from previous layers. Therefore, following~\cite{IFQNet}, we employ  the widely used one-stage pipeline YOLOv2 for our face detector.


\subsection{Deep Network Compression}

To reduce the computation cost of the deep models, many  approaches have been studied. One way is to design novel efficient architectures. For example, by replacing a standard convolution layer by the combination of a depth-wise and a point-wise (1$\times$1) convolution layer, MobileNets~\cite{MobileNetv1} reduces the weights and computation by 8$\times$$\sim$9$\times$ times. Similarly, LBCNN~\cite{LBCNN} employs predefined binary patterns for the depthwise convolution and shares those patterns over multiple layers for further compression.

Another direction for designing a compact model is to compress the network through pruning, quantization, etc. Pruning methods eliminate the less important connections and fine-tune the pruned network to narrow down the accuracy drop. For example, in~\cite{QunatMimic}, Wei~\etal reduce the input and output channels of each layer of VGG by 32 times and design a  very small detector whose size is only 132KB. In contrast, quantization approaches aim to quantize the float data of a network into low-bits data. For example, XNOR-Net~\cite{XNOR} and HWGQ-Net~\cite{HWGQ} achieves 32$\times$ times savings on model size via binarizing (1-bit) the network weights. In addition to the quantized weights,  further quantizing feature maps into low-bits data can reduce the feature maps memory usage and meanwhile increase the inference speed. For example, XNOR-Net which quantizes both weights and feature maps into 1-bit is theoretically 64$\times$ times faster than its full precision counterpart. Furthermore, for embedded devices such as FPGA and ASIC, quantization network is particularly attractive because it leads to higher throughput and lower power consumption through converting the network into a fixed-point one.  

One interesting topic is about further exploring the redundancy and  compressing the quantized network. For example, in~\cite{ToCompress}, various networks (VGG16, MobileNet) are firstly quantized to 8-bit data and then further pruned by 24\%. Similarly, Li and Ren~\cite{BNNPrune} explores the redundancy of a Binarized Neural Network (BNN) and further compresses the model size by 3.9$\times$ times through bit-level data pruning. 

Different with methods that explore the redundancy through carefully tuned strategies, we propose to simply employ duplicated weights which contain lots of identical channels for the weight-intensive layers. Since the duplicated weights can be easily restored from the template weights which contain all the non-identical channels, it is sufficient to only store the template weights in the memory during inference time.

\subsection{Accuracy Improvement for Quantized Network}

Even though quantizing the network into low-bits data leads to promising reduction on computation cost, accuracy drop is usually observed. As demonstrated in~\cite{Google8bit,NvidiaInt8}, quantizing the network data into 8-bits  only leads to minor accuracy drop on ImageNet classification task~\cite{ImageNet}. Nevertheless, quantizing the network into lower bits usually results in notable accuracy degradation.  For example, XNOR-Net~\cite{XNOR} which quantizes both its weights and feature maps into 1-bit and thus observes a 12.6\% accuracy drop (56.8\% vs. 44.2\%). Based on that, HWGQ-Net~\cite{HWGQ} gains 8.2\% accuracy back through using 2-bits on its feature maps (52.4\% on ImageNet). 
Additionally, for object detection tasks, 3\%$\sim$5\% drop is observed in~\cite{SQuantizer}.

To improve the accuracy of quantized networks, lots of efforts have been done on better strategies for training the networks. INQ~\cite{INQ} proposes to incrementally quantize the weights and achieves more accurate quantized networks through iterative fine-tuning. Similarly, in~\cite{PQ}, the weights and activations are firstly quantized to 16-bits, then to 4-bits and at the end to 2-bits. PACT~\cite{PACT} optimizes the clipping thresholds for better quantization on feature maps. Besides, knowledge distillation technology additionally uses the knowledge from teacher network to guide the training process of student network~\cite{QunatMimic}. 

To narrow down the accuracy drop caused by the network quantization, we propose to duplicate the feature maps of its quantization-sensitive layers  which allows us to use weights with more channels for convolving more representative features. The advantage of our method is that it gives significant accuracy improvement without increasing the feature maps memory usage.

\section{Our Approach: DupNet}\label{Sec:hwgq}
To further compress the model size and improve the accuracy of a quantized network for face detection, we propose to employ weights with duplicated channels in the weight-intensive layers and duplicate the input feature maps of its quantization-sensitive layers. 

\begin{figure*}[!htb]
	\centering
	\includegraphics[width=\linewidth]{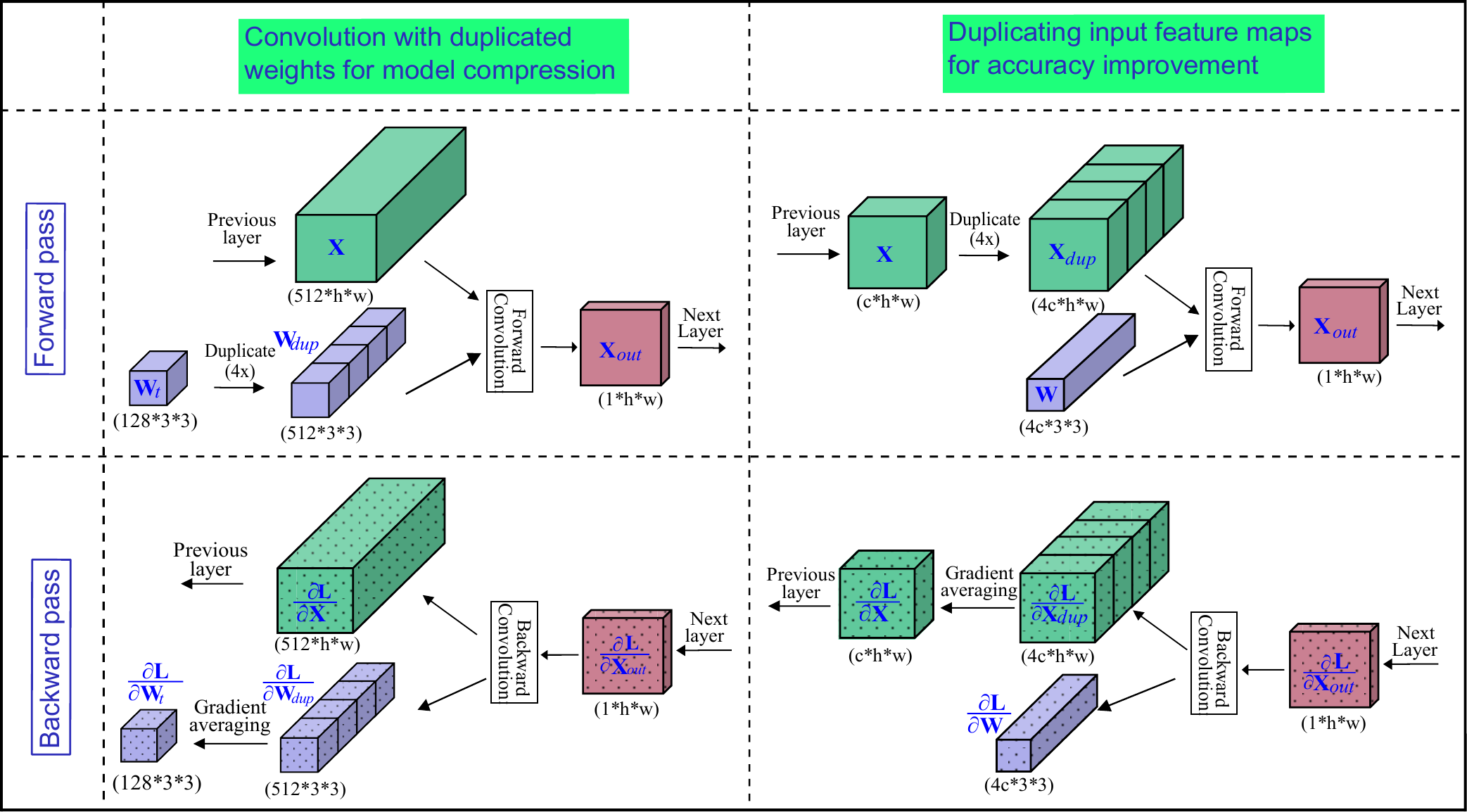}
	
	\caption{Illustration of the forward and backward pass for both duplicated weights and feature maps duplication.}
	\label{fig:WXdup}
\end{figure*}

\subsection{Duplicated Weights for Model Compression}\label{subSec:Wdup}
As discussed in~\cite{BNNPrune}, even though the network data is quantized into very low-bits data, redundancy still exists. In this section, we will illustrate our method which employs template weights with less channels and thus less redundancy. During convolution process, we duplicate the template weights to get required channels to convolve with the input feature maps.

We assume the quantized network only employs $a2w1$ convolution which means that the input feature maps and weights are quantized to 2-bits and 1-bit respectively. We represent its weights and feature maps as $\textbf{W}\in\{-1, +1\}^{c\times 3 \times 3}$ and $\textbf{X}\in \{0, 1, 2, 3\}^{c\times h \times w}$, where $c$, $h$, $w$ are the number of channels, the width and height of input feature maps, $3 \times 3$ is the kernel size of the convolution\footnote{Even though we use $3 \times 3$ for explanation, our method is also able to compress the convolution with other kernel size (e.g. $1 \times 1$ and $5 \times 5$). }. 
To compress the model size, we define a weights template $\textbf{W}_t\in\mathbb{R}^{c{'}\times 3 \times 3}$ which has less channels ($c{'} < c$). However, it can not be used to convolve with $\textbf{X}$ since they have different number of channels. To solve such problem, we duplicate the channels of template weights into required number and obtain duplicated weights $\textbf{W}_{dup} \in\mathbb{R}^{c\times 3 \times 3}$.

During training process, it is straightforward to compute the gradient of duplicated weights $\frac{\partial L}{\partial \textbf{W}_{dup}}$ by employing standard convolution, where $L$ represents the loss of the network for given training samples. As shown in Figure~\ref{fig:WXdup}, to compute the gradient of template weights $\frac{\partial L}{\partial \textbf{W}_{t}}$, we average the corresponding channels of the $\frac{\partial L}{\partial \textbf{W}_{dup}}$. Specifically, we assume that $c= 512$ and $c{'}=128$ for 4$\times$ times compression, and the 0th, 128th, 256th and 384th channels of $\textbf{W}_{dup}$ are duplicated from the 0th channel of $\textbf{W}_{t}$.  In order to compute the 0th channel of the gradient $\frac{\partial L}{\partial \textbf{W}_{t}}$, we element-wisely average the 0th, 128th, 256th and 384th channels of $\frac{\partial L}{\partial \textbf{W}_{dup}}$. Such gradient averaging process is repeated for computing the 1$\sim$127th channels of $\frac{\partial L}{\partial \textbf{W}_{t}}$. At the end, the template weights $\textbf{W}_{t}$ can be learned by iteratively updating it with its gradients $\frac{\partial L}{\partial \textbf{W}_{t}}$ using SGD optimization.

Since the duplicated weights $\textbf{W}_{dup}$ can be easily restored from the templates weights $\textbf{W}_t$, it is only necessary to store $\textbf{W}_t$. Thus,  $c/c{'}$ times model size reduction can be achieved. Nevertheless, our compression method may harm the accuracy because the duplicated weights contains less non-identical channels. Considering that the model size is usually dominated by the weight-intensive layers, we only apply our compression method on these layers to prevent significant accuracy drop.

Furthermore, given the fact that many channels of the duplicated weights are identical, we can reduce the computation complexity as follows. We split duplicated weights $\textbf{W}_{dup}$ into $\textbf{W}_1, \textbf{W}_2, \textbf{W}_3, \textbf{W}_4$ while each of them is identical with the template weights $\textbf{W}_t$. Similarly, the feature maps $\textbf{X}$ can also be accordingly split into $\textbf{X}_1, \textbf{X}_2, \textbf{X}_3, \textbf{X}_4$ which are non-identical. We use  $\otimes$  and $Concat(,)$ to represent convolution operation and a function that concatenate its members along channel axis respectively. Then, $\textbf{W}_{dup} \otimes \textbf{X} = Concat(\textbf{W}_1, \textbf{W}_2, \textbf{W}_3, \textbf{W}_4) \otimes Concat(\textbf{X}_1, \textbf{X}_2, \textbf{X}_3, \textbf{X}_4) = \textbf{W}_1 \otimes \textbf{X}_1 + \textbf{W}_2 \otimes \textbf{X}_2 + \textbf{W}_3 \otimes \textbf{X}_3 + \textbf{W}_4 \otimes \textbf{X}_4 = \textbf{W}_{t} \otimes (\textbf{X}_1 + \textbf{X}_2 + \textbf{X}_3 + \textbf{X}_4)$. Consequently, the convolution $\textbf{W}_{dup} \otimes \textbf{X}$ can be alternatively computed by $\textbf{W}_{t} \otimes \textbf{X}_{sum}$,  where $\textbf{X}_{sum} =\textbf{X}_1 + \textbf{X}_2 + \textbf{X}_3 + \textbf{X}_4$.  The overall computation complexity of $\textbf{W}_{t} \otimes \textbf{X}_{sum}$ and $\textbf{X}_{sum}$ is much smaller than $\textbf{W}_{dup} \otimes \textbf{X}$.

\subsection{Duplicate Feature Maps to Improve Accuracy}\label{subSec:Xdup}

The quantized networks quantize their full precision data into low-bits data thus usually leads to notable accuracy drop. In the following, we further improve the degraded accuracy for a very tiny quantized face detector.

As discussed in Section~\ref{sec:intro}, the accuracy degradation of quantized face detector is mainly caused by the weak representation power of the quantized output feature maps of its quantization-sensitive layers. To enhance the representative power of their output features, one straightforward way is to simply make these layers wider (more input feature maps). However, it significantly increases the memory usage on feature maps.  Besides, observing that these layers usually locate in the lower part of a network, such memory increase may cause critical issue because their feature maps hold high resolution ($h \times w$). In contrast, the number of channels ($c$) is usually small and thus their weights ($c \times 3 \times 3$) is small too. Consequently, we propose to  duplicate the input feature maps and employ more weights channels for better output features. As shown in Figure~\ref{fig:WXdup}, the input feature maps are duplicated 4$\times$ times and thus the weights size is also increased to $4c \times 3 \times 3$. For the backward pass during training time, to obtain the gradients $ \frac{\partial L}{\partial \textbf{X}}$, we firstly compute the gradients of duplicated feature maps $\frac{\partial L}{\partial \textbf{X}_{dup}}$, and then average every 4 of its corresponding channels that are identical in $\textbf{X}_{dup}$. At the end, the gradients $ \frac{\partial L}{\partial \textbf{X}}$  are propagated back to its previous layers.

Comparing  with the strategy that simply uses more input feature maps, our method does not require extra memory for feature maps. Thanks to the increased channels of input feature maps, we can employ more channels of weights ($c$ vs. $4c$). Consequently, the input feature maps are convolved with more patterns and thus leads to more representative power on the resulted features. Even though our method increases the weights size, such cost increase has limited influence on the overall model cost because the weights size of these layers are usually very small (see Table~\ref{tab:IFQ_cost}). Similar with the theory that is explained in Section~\ref{subSec:Wdup}, one also can achieve speedup by replacing  $\textbf{W} \otimes \textbf{X}_{dup}$ with $\textbf{W}_{sum} \otimes \textbf{X}$,  where $\textbf{W}_{sum}$ can be obtained by summing the corresponding channels of $\textbf{W}$.

\section{Experimental Results}

To design a very tiny CNN for face detection, we borrow the compression ideas from IFQ-Tinier-YOLO~\cite{IFQNet} which compresses Tiny-YOLO network by 260$\times$ times through halving the filter numbers of all convolution layers, replacing one 3$\times$3 layer which contains massive parameters by 1$\times$1 kernels and binarizing the weights in all layers. Moreover, we further halve their filter number and apply the proposed duplicated weights for its weight-intensive layers (Conv6$\sim$Conv8) and achieve 6.7$\times$ times further savings on model size. Besides, we will demonstrate that duplicating the input feature maps of its quantization-sensitive layers can significantly improve the accuracy.

We employ WiderFace~\cite{widerface} training images to train our models using Darknet framework~\cite{darknet}. For fair comparison, all the models are trained with the same strategies which are: 1) training the models by 100k iterations with SGD optimization method; 2) the learning rate is initially set to 0.01 and downscaled by a factor of 0.1 at the $30k$-th, $60k$-th, $80k$-th and $90k$-th iteration respectively; 3) all the models are trained from scratch. Furthermore, we use FDDB~\cite{FDDB} benchmark which contains 5,171 faces within 2,845 test images to evaluate the accuracy of our face detectors. Inheriting from~\cite{IFQNet}, we use the detection rate when 284 false positive faces are reached (averagely allowing 1 false positive in every 10 images) as the evaluation metric.

\subsection{Model Compression} \label{subsec:exp-Wdup}

To further compress model size of IFQ-Tinier-YOLO, we analyze the weights size for each of its layer. Meanwhile, to measure the computation complexity, we borrow the term \#FLOPs\footnote{As stated in ~\cite{XNOR}, for the 64-bit based computing devices, 64 Multi-Adds of the $a1w1$ convolution are equivalent to 1 FLOP. Similarly, we assume that 32 Multi-Adds of the $a2w1$ convolution and 8 Multi-Adds of the $a8w1$ (Conv1) equal to 1 FLOP respectively.} (Floating-point operations) which is generally used for full precision networks~\cite{FLOPs}. Nevertheless, it is worthy to point out that our network can be lossless converted to fixed-point network and thus does not require any floating-point operation.

\begin{table}[!h]
	\centering
	\caption{IFQ-Tinier-YOLO inference costs in terms of weights size and  \#FLOPs (million) for measuring computation complexity.}
	\label{tab:IFQ_cost}
	\setlength\tabcolsep{1.5pt}
	\begin{tabular}{c|c|c|c|c}
		
		\hline
		\multirow{2}{*}{} &\multirow{2}{*}{\shortstack{Kernel \\size (W)}} & \multirow{2}{*}{\shortstack{Feature \\size (X)}}  &\multirow{2}{*}{\shortstack{\#FLOPs\\(million)}}  &\multirow{2}{*}{\shortstack{Weights \\size (KB)}}\\
		& & & & \\
		\hline \hline
		
		Conv1     &$8 \times 3 \times 3$      &$608 \times 608$            &10.0              &$<0.1$ \\
		Conv2     &$16 \times 3 \times 3$     &$304 \times 304$            &3.3               &0.1 \\
		Conv3     &$32 \times 3 \times 3$     &$152 \times 152$            &3.3               &0.6  \\
		Conv4     &$64 \times 3 \times 3$     &$76  \times 76$             &3.3               &2.3  \\
		Conv5     &$128 \times 3 \times 3$    &$38  \times 38$             &3.3               &9 \\
		Conv6     &$256 \times 3 \times 3$    &$38  \times 38$             &13.3               &36 \\
		Conv7     &$512 \times 3 \times 3$    &$38  \times 38$             &53.2              &144 \\
		Conv8     &$512 \times 1 \times 1$    &$38  \times 38$             &11.8               &32 \\
		Conv9     &$30 \times 3 \times 3$     &$38  \times 38$             &6.2               &8.44 \\
		\hline
		\multicolumn{3}{c|}{Overall}           &107.9              &240.9 \\	
		
		\hline
	\end{tabular}
\end{table}

As shown in Table~\ref{tab:IFQ_cost}, the weight-intensive layers of IFQ-Tinier-YOLO model are Conv6$\sim$Conv8 layers. Consequently, to further compress the model size, we apply two techniques for these layers: halving the filters number and employing duplicated weights. Regarding to the duplicated weights, we casted experiments that employ 2$\times$ or 4$\times$ or 8$\times$ times duplication to figure out the optimal trade-off between high compression ratio and low detection rate. 

\begin{figure}[!htb]
	\begin{center}
		\includegraphics[width=0.8\linewidth]{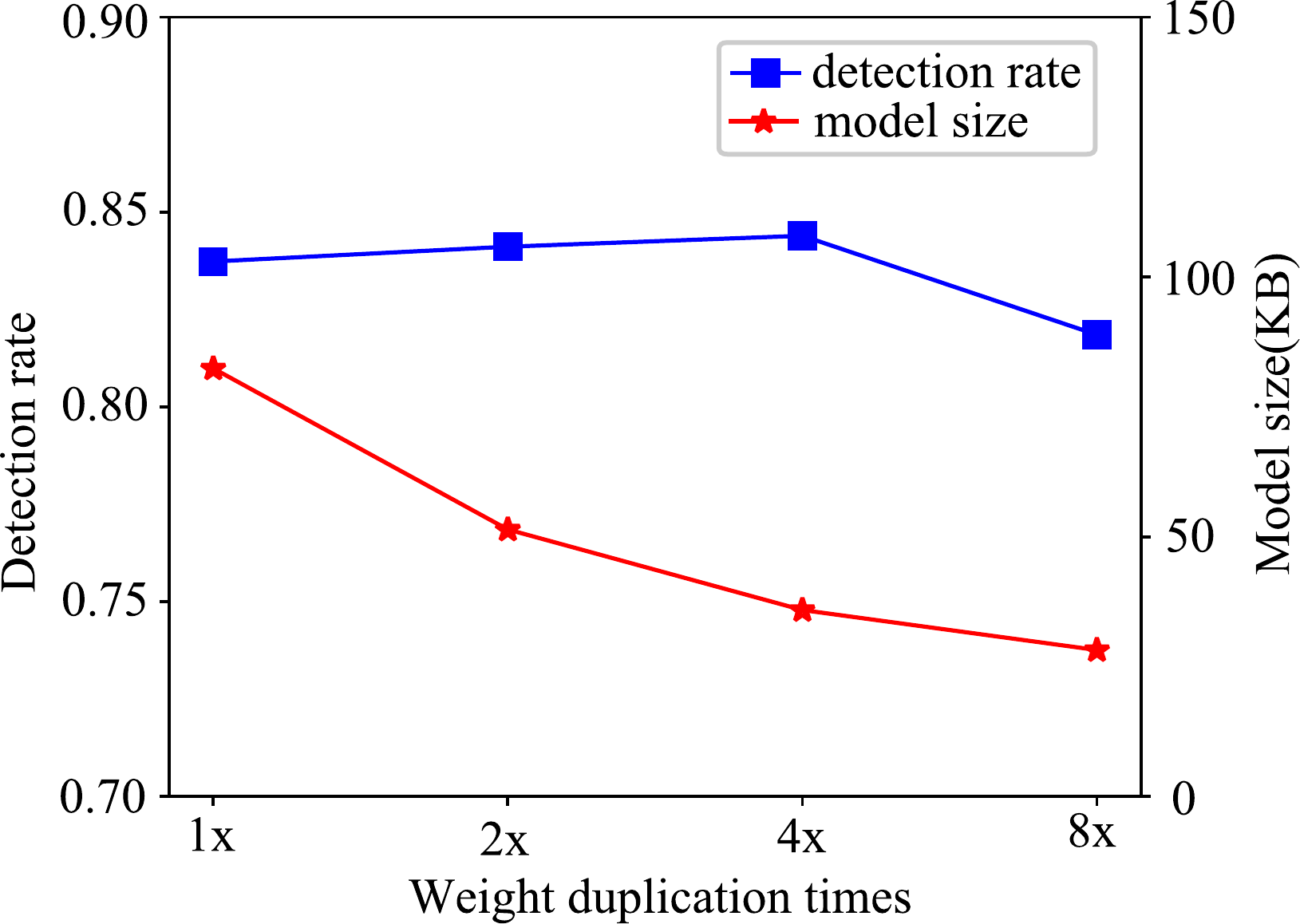}
	\end{center}
	\caption{Comparison of the detection rate for the models with various compression ratio on Conv6$\sim$Conv8 layers.}
	\label{fig:exp_W_dup}
\end{figure}

We first halve the filter number of the weight-intensive layers and thus reduce its model size from 240.9 KB to 82.4 KB (marked as ``1$\times$" in Figure~\ref{fig:exp_W_dup}). Meanwhile, it achieves 0.837 on detection rate which is very close to IFQ-Tinier-YOLO (0.84~\cite{IFQNet}). Additionally, as shown in Figure~\ref{fig:exp_W_dup}, employing 2$\times$ or 4$\times$ times duplicated weights gives further reduction on model size without detection rate drop. More specifically, with the help of halved filter number and 4$\times$ times duplicated weights, we reduce the model size of IFQ-Tinier-YOLO from 240.9 KB to 35.9 KB indicating a 6.7$\times$ times reduction. Furthermore,  when compressing the Conv6$\sim$Conv8 by 8$\times$ times, the accuracy only decreases by 2.1\% while the compression ratio increases to 8.5$\times$ times. 

\begin{figure}[!htb]
	\begin{center}
		\includegraphics[width=0.8\linewidth]{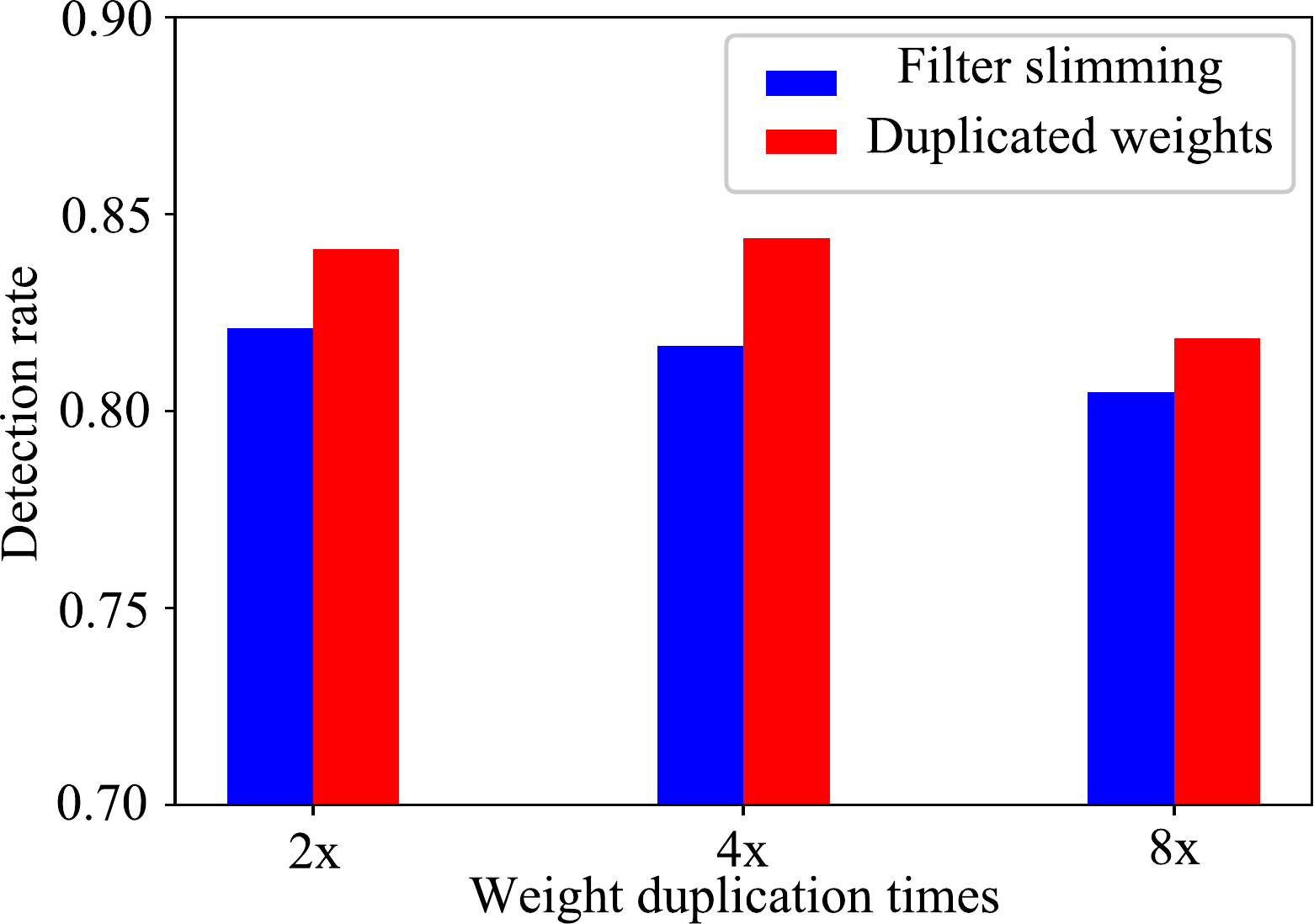}
	\end{center}
	\caption{Performance of our duplicated weights based method and the filter slimming method for model compression.}
	\label{fig:exp_compress_compare}
\end{figure}

The reason that our compression method does not give notable accuracy drop is that the redundant connections exist in those three layers. However, one may argue that further reducing the number of their filters also can reduce the model size. Consequently, we compare such method (marked as ``Filter slimming") with our method in Figure~\ref{fig:exp_compress_compare}. For fair comparison, we reduce the filter numbers of Conv6$\sim$Conv8 to make them have similar model size with our duplicated weights models. For example, to compare with our model with 4$\times$ times compression on all Conv6$\sim$Conv8 layers, we instead halve their filter numbers resulting in a 2$\times$, 4$\times$ and 2$\times$ times compression for these three layers respectively. As demonstrated in Figure~\ref{fig:exp_compress_compare}, for different compression ratios, our weights duplication based method generally outperforms the filter slimming method. 

In the above, we have demonstrated that our duplicated weights based compression is very effective for the quantized network whose precision is $a2w1$. To demonstrate the generalization ability of our method, we further test it on the networks that are quantized into different precisions. As shown in Table~\ref{tab:W_dup}, our method with 4$\times$ times weights duplication also gives no accuracy drop for the $a2w2$, $a4w4$ and $a8w8$ networks. When further compressing them by 8$\times$ times, slight degradation is observed. One interesting observation is that the higher precision the network is, the less accuracy drop is caused. For example, with 8$\times$ times compression, the detection rate drop for $a2w2$ network is 3.8\% while it is only 2.0\% for $a4w4$ and 0.6\% for $a8w8$. We think it is because that the more accurate the network is, the more redundancy usually exists in its connections.

\begin{table}[!h]
	\centering
	\caption{Performance of our compression method on the face detectors with various quantization precision.}
	\label{tab:W_dup}
	\begin{tabular}{c|c|c|c|c}
		\hline
		
		\multirow{2}{*}{\shortstack{Weights \\duplication}} &\multicolumn{4}{c}{Network precision}\\
		\cline{2-5}
		&$a2w1$            &$a2w2$              &$a4w4$            &$a8w8$ \\
		\hline\hline
		{1$\times$}    &0.837             &0.866               &0.892             &0.906  \\
		{2$\times$}    &0.841             &0.862               &0.888             &0.907  \\
		{4$\times$}    &0.844             &0.865               &0.890             &0.900\\
		{8$\times$}    &0.819             &0.828               &0.872             &0.892\\
		\hline
	\end{tabular}
\end{table}

\subsection{Accuracy Improvement} \label{subsec:exp-Xdup}

The accuracy of quantized networks usually is notably lower than their full precision counterparts. For example, the quantized network based face detector, IFQ-Tinier-YOLO leads to $\sim$6\% drop on detection rate~\cite{IFQNet}. On the other hand, quantizing different layers leads to  widely-varied performance loss~\cite{LWB}. To improve the accuracy, we first locate the quantization-sensitive layers of a quantized face detector through layer-wise quantization strategy.

\begin{table}[!h]
	\centering
	\caption{Layer-wise quantization to locate the source of accuracy drop.}
	\label{tab:exp_lwb}
	\setlength\tabcolsep{1.5pt}
	\begin{tabular}{c|c|c|c|c|c|c}
		
		\hline
		\multicolumn{4}{c|}{Quantized Conv. layers}  &\multirow{2}{*}{\shortstack{\#FLOPs\\(million)}} &\multirow{2}{*}{\shortstack{Model \\size(KB)}} &\multirow{2}{*}{\shortstack{Detection \\rate}}\\
		\cline{1-4}
		{1st}    &{2nd-3rd}       &{4th-8th} &{9th} &  &  &  \\
		
		\hline \hline		
		&              &              &               &1,338.9    &2,637.3   &0.902\\
		&              &$\checkmark$  &               &422.2      &366.6     &0.880\\
		&$\checkmark$  &$\checkmark$  &               &215.9      &344.8     &0.858\\
		$\checkmark$  &$\checkmark$  &$\checkmark$  &               &146.0      &344.0     &0.845\\
		$\checkmark$  &$\checkmark$  &$\checkmark$  &$\checkmark$   &49.3       &82.4      &0.837\\
		\hline
	\end{tabular}
\end{table}

As shown in Table~\ref{tab:exp_lwb}, we firstly quantize Conv4$\sim$Conv8 convolution layers of a full precision counterpart of IFQ-Tinier-YOLO network but with halved filter number in Conv6$\sim$Conv8. In this subsection, to demonstrate the accuracy improvement effect of duplicating the input feature maps, the duplicated weights based compression is not applied. As shown in Table~\ref{tab:exp_lwb}, quantizing Conv4$\sim$Conv8 leads to 3.2$\times$ and 7.2$\times$ times reduction on MFLOPs (million of FLOPs) and model size respectively while the detection rate only drops by 2.2\%. Nevertheless, progressively quantizing the Conv3$\sim$Conv2 and then Conv1 causes  2.2\%, 1.3\% accuracy drop respectively but gives much less reductions on inference cost.   
Thus, we define the Conv1$\sim$Conv3 as quantization-sensitive layers of the network. We think the reason is that they only contain limited number of feature maps. Consequently, quantizing them severely damages the representative power of their output features. At the end, quantizing Conv9, resulting in a fully quantized Tinier-YOLO model, further gives remarkable savings on computation cost while the accuracy is only decreased by 0.8\%. 

\begin{table}[!h]
	\centering
	\caption{Illustration of performance improvement for progressively duplicating the input feature maps of the Conv2-3, Conv1 and Conv9 of fully quantized Tinier-YOLO face detector.}
	\label{tab:exp_X_dup}
	\setlength\tabcolsep{1.5pt}
	\begin{tabular}{c|c|c|c|c|c|c}
		
		\hline
		\multicolumn{4}{c|}{Feature maps duplication?} &\multirow{2}{*}{\shortstack{\#FLOPs\\(million)}} &\multirow{2}{*}{\shortstack{Model \\size(KB)}} &\multirow{2}{*}{\shortstack{Detection \\rate}}\\
		\cline{1-4}
		{Conv1}    &{Conv2}      &{Conv3}   &{Conv9} &  &  & \\
		
		\hline \hline
		
		&              &              &              &49.3             &82.4       &0.837\\
		&$\checkmark$  &$\checkmark$  &              &62.6             &83.4       &0.867\\
		$\checkmark$  &$\checkmark$  &$\checkmark$  &              &92.6             &83.5       &0.872\\
		$\checkmark$  &$\checkmark$  &$\checkmark$  &$\checkmark$  &95.7             &91.9       &\textbf{0.890}\\
		\hline
	\end{tabular}
\end{table}

To improve the accuracy of the fully quantized Tinier-YOLO, we firstly duplicate the input feature maps of Conv2 and Conv3 by 4$\times$ and 2$\times$ times respectively. As shown in Table~\ref{tab:exp_X_dup}, it gives 3.0\% increase on detection rate while the model size and computation complexity are only increased by 1.2\% and 27.0\% respectively. 
Furthermore, additionally duplicating the feature maps of Conv1 by 4$\times$ times gives 0.5\% increase on detection rate while the model size only increases 0.1KB. However, its computation complexity is increased from 62.6 MFLOPs to 92.6 MFLOPs (47.9\% increase).  At the end, we further duplicate the feature maps of Conv9 by 2x and achieve 1.8\% improvement on detection rate at the price of 3.3\% and 10.1\% increase on \#FLOPs and model size respectively.

\begin{table}[!h]
	\centering
	\caption{Comparison between our method and the quantization precision increasing method on improving the detection rate.}
	\label{tab:exp_improv_comp}
	\begin{tabular}{c|c|c|c}
		
		\hline
		\multicolumn{3}{c|}{Quantization precision} &\multirow{2}{*}{Detection rate} \\
		\cline{1-3}
		{Conv1} & {Conv2}            & {Conv3}              & \\
		\hline\hline
		{$a8w1$}     &{$a2w1$(4$\times$)}            &{$a2w1$(2$\times$)}        &\textbf{0.867}          \\
		{$a8w1$}     &{$a2w3$}                &{$a2w2$}            &{0.850}          \\
		\hline
		{$a8w1$(4$\times$)} &{$a2w1$(4$\times$)}            &{$a2w1$(2$\times$)}        &\textbf{0.872}          \\
		{$a8w3$}     &{$a2w3$}                &{$a2w2$}            &{0.860}          \\
		
		\hline	
	\end{tabular}
\end{table}

On the other hand, employing more bits for the weights of quantization-sensitive layers can also improve the accuracy. For fair comparison, in the case of 4$\times$ times duplication (\eg Conv2), we use 3-bits on weights (``$a2w3$'') to compare it with our method (``$a2w1(4\times)$'') which can be computed by $W_{sum}\otimes X$ where $W_{sum}\in \{-4,-2,0,2,4\}$\footnote{Each elements of  $W_{sum}$ is the summation of four binary elements (either -1 or +1) from four corresponding channels (see Section~\ref{subSec:Xdup}).} that can be represented using 3-bits.  As shown in Table~\ref{tab:exp_improv_comp}, our methods generally gives more than $1.0\%$ improvement on detection rate. Furthermore, our method is more attractive for hardware design in three aspects: 1) it use less information (only 5 values vs. 8 values) which makes the coding-based further compression easier (\eg Huffman coding~\cite{DeepCompression} and RLC~\cite{RLC}); 2) lots of its weights are 0 thus the corresponding computation can be optimized; 3) our model can be computed only using  $a2w1$ convolution\footnote{The $a8w1$ convolution (Conv1) can be computed by the accumulation of four $a2w1$ convolutions.} which can make the hardware design simpler.

\subsection{Face Detectors Comparison} \label{subsec:exp-comp}

As demonstrated in the previous experiments, employing duplicated weights gives remarkable compression without obvious accuracy drop. On the other hand, duplicating the feature maps for the quantization-sensitive layers improves the detection rate by a large margin. In this section, we combine these two technique to design DupNet-Tinier-YOLO which is a very tiny quantized face detector with improved accuracy. In details, we employ 4$\times$ times compression for weight-intensive layers (Conv6$\sim$Conv8) and duplicate the input feature maps of Conv2$\sim$Conv3. We initially choose not to duplicate the input feature maps of Conv1 in DupNet-Tinier-YOLO to avoid notable increase on \#FLOPs. Regarding to the model size, 4$\times$ times weights compression reduces the model size from 82.4KB to 35.9 KB and duplicating the input feature maps increase it to 36.9 KB.

%

\begin{table}[!h]
	\centering
	\caption{Comparison of the face detectors in terms of the computation complexity (\#FLOPs), model size and detection rate.}
	\label{tab:exp_alg_compare}
	\setlength\tabcolsep{1.5pt}
	\begin{tabular}{c|c|c|c}
		
		\hline      	
		{Models}    &{\shortstack{\#FLOPs\\(million)}}     &{\shortstack{Model \\size(KB)}}   &\shortstack{Detection\\ rate} \\
		
		\hline \hline
		
		Tiny-YOLO                     &24,407         &62,516            &0.920            \\
		Tinier-YOLO                   &3,213          &7,707             &0.902            \\
		IFQ-Tinier-YOLO~\cite{IFQNet} &107.9          &240.9             &0.835           \\
		DupNet                       &62.6           &36.9              &{0.859}  \\
		DupNet+PACT                  &62.6           &36.9              &{0.880}  \\
		DupNet-L                       &95.7           &45.4              &{0.884}  \\
		DupNet-L+PACT                  &95.7           &45.4              &{0.906}  \\

		
		\hline
	\end{tabular}
\end{table}

\begin{figure}[!htb]
	\begin{center}
		\includegraphics[width=0.9\linewidth]{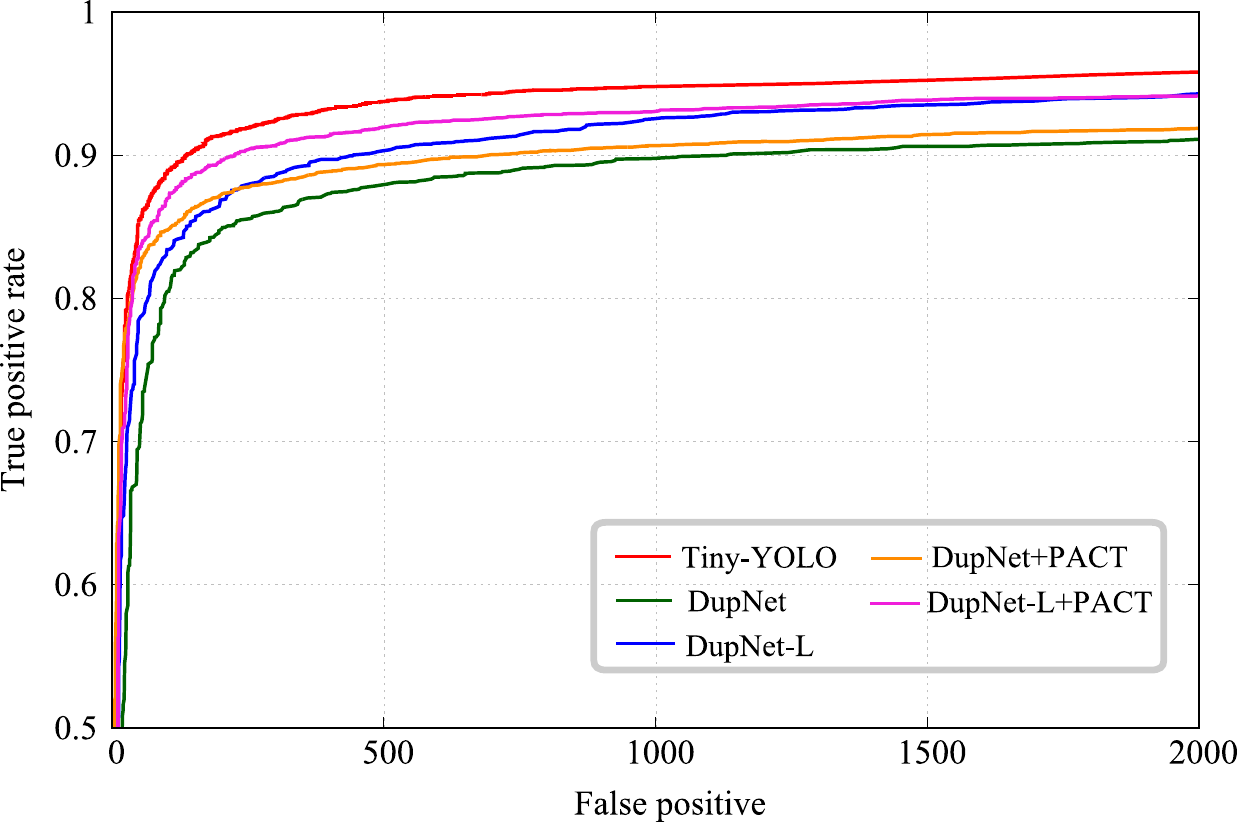}
	\end{center}
	\caption{Comparison on the performance of the face detectors  in terms of ROC curves of FDDB dataset~\cite{FDDB}.}
	\label{fig:exp_alg_comp}
\end{figure}

\begin{figure*}[!htb]
	\centering
	\includegraphics[width=\linewidth]{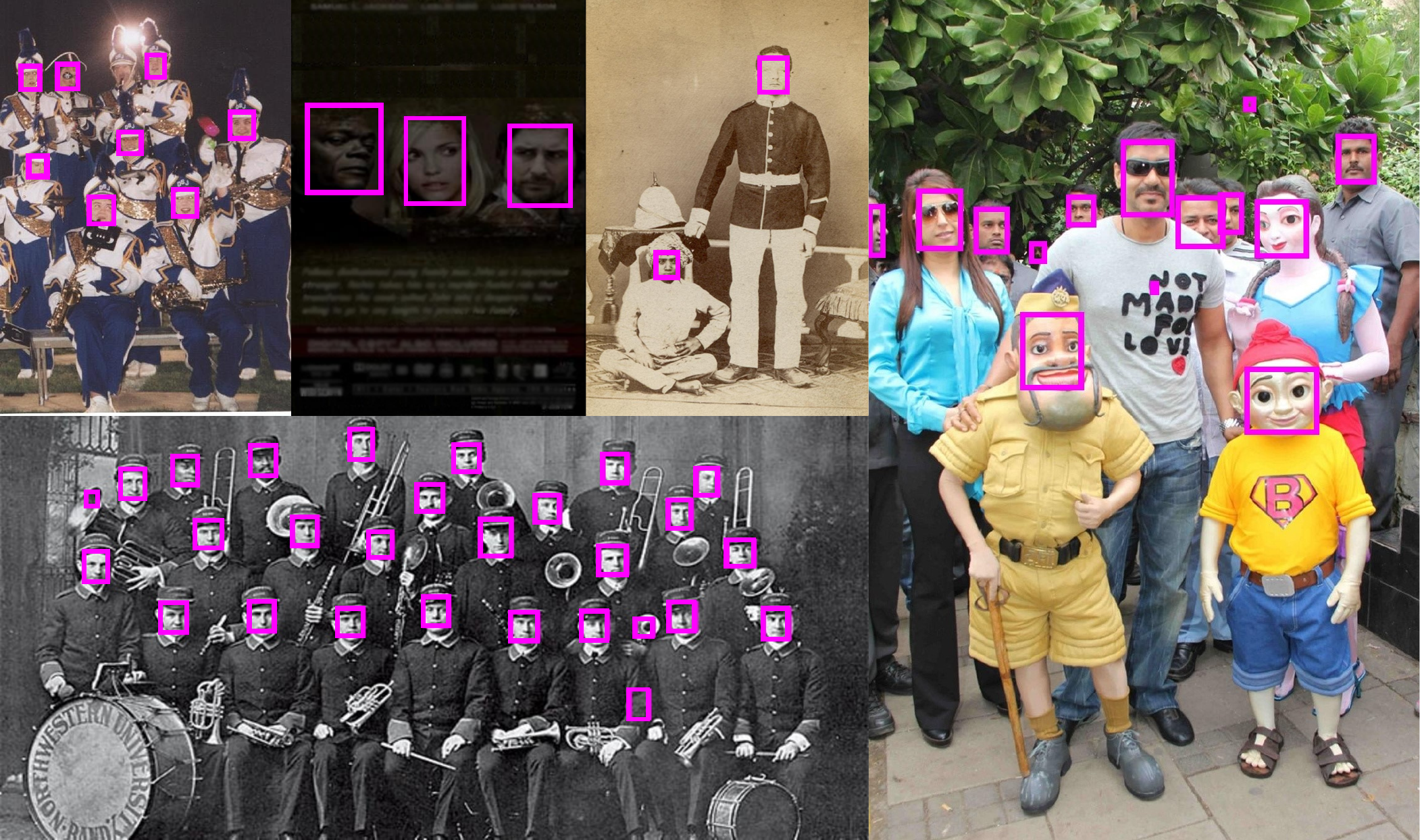}
	\caption{Qualitative results of the proposed DupNet-Tinier-YOLO-L face detector on Wider Face dataset~\cite{widerface}. }
	\label{fig:exp_widerFace}
\end{figure*}

As shown in Table~\ref{tab:exp_alg_compare}, comparing with the IFQ-Tinier-YOLO~\cite{IFQNet}, our DupNet-Tinier-YOLO (represented as ``DupNet") gives 6.5$\times$ times savings on model size and 42.0\%  
less MFLOPs. Meanwhile, it also gives 2.4\% improvement on detection rate. To further improve the detection rate with acceptable cost increase, we design DupNet-Tinier-YOLO-L (marked as ``DupNet-L")  which additionally duplicates the input feature maps fo Conv1 and Conv9 by 4$\times$ and 2$\times$ respectively. As shown in Table~\ref{tab:exp_alg_compare},  DupNet-Tinier-YOLO-L further gives 2.5\% higher detection rate. Nevertheless, the model size and \#FLOPs are increased to 45.4 KB and 95.7 MFLOPs respectively, both of which are still smaller than IFQ-Tinier-YOLO.

Furthermore, we employ PACT~\cite{PACT} to train optimal clipping thresholds for feature maps quantization to improve the accuracy. As illustrated in Table~\ref{tab:exp_alg_compare}, PACT algorithm improves the DupNet-Tinier-YOLO and DupNet-Tinier-YOLO-L by 2.1\% and 2.2\% respectively. Comparing with Tiny-YOLO network, our DupNet-Tinier-YOLO achieves 389.9$\times$ and 1694.2$\times$ times reduction on \#FLOPs and model size respectively while the detection rate is only decreased by 4.0\%. On the other hand, the accuracy of DupNet-Tinier-YOLO-L is only decreased by 1.4\% while the inference cost reduction is kept impressive. Besides, we compare their accuracy in terms of ROC curves in Figure~\ref{fig:exp_alg_comp}.

To demonstrate the performance of our detector on more challenging faces, we also test DupNet-Tinier-YOLO-L on WiderFace testing dataset~\cite{widerface}. As shown in Figure~\ref{fig:exp_widerFace}, our model also gives excellent detection quality in various challenging scenarios such as tiny size, low-illumination, severe occlusion and degraded coloring, etc.

In summary, we proposed DupNet-Tinier-YOLO face detector, which is quantized, very tiny and accurate. By employing duplicated weights for the weight-intensive layers, we reduced the model size and \#FLOPs of IFQ-Tinier-YOLO by 6.5$\times$ times and 42.0\% respectively. Meanwhile, we increased its detection rate by 4.5\% by using the proposed DupNet and the PACT~\cite{PACT} technique.  Moreover, we demonstrated that our DupNet can be flexibly adjusted for different inference cost (\eg DupNet-Tinier-YOLO-L has higher cost and is more accurate).

\section{Conclusions}
In this paper, we proposed DupNet which employs duplicated weights for the weight-intensive layers of a quantized CNN to compress its model size. Furthermore, we observe that the degraded accuracy of the quantized CNN is mainly caused by quantization-sensitive layers which have poor representative power on their quantized output feature maps. Hence, DupNet also duplicates the input feature maps of these layers and employ more weights channels to improve their output features. Through the experiments on FDDB dataset, we demonstrated that our DupNet-Tinier-YOLO face detector can significantly compress the model size and meanwhile impressively improve the detection rate. Moreover, our DupNet-Tinier-YOLO face detector can be lossless converted into fixed-point network~\cite{IFQNet} and thus can be easily implemented on embedded devices.

Additionally, our DupNet can be combined with other algorithms that are proposed to improve the performance of compressed networks such as knowledge distillation. Moreover, despite we only test our method on face detection, it is also applicable for other tasks such as object detection or even face recognition or semantic segmentation, etc.

{\small
\bibliographystyle{ieee_fullname}
\bibliography{paper_HG}
}

\end{document}